\documentclass[letterpaper, 10 pt, conference]{ieeeconf}  

\IEEEoverridecommandlockouts                              

\usepackage[hidelinks]{hyperref} 
\usepackage{graphics} 
\usepackage{amsmath} 
\usepackage{amssymb}  
\usepackage{theorem}
\usepackage{multirow}
\usepackage{xcolor}
\usepackage{algorithm}
\usepackage{algpseudocode}
\usepackage{algorithmicx}
\usepackage{graphicx}

\usepackage{tikz}
\usepackage{textcomp}
\usepackage{hyperref}
\usepackage{tabularx}
\usepackage{booktabs}

\newcommand\copyrighttext{%
  \footnotesize This work has been submitted to the IEEE for possible publication. Copyright may be transferred without notice, after which this version may no longer be accessible.}
\newcommand\copyrightnotice{%
\begin{tikzpicture}[remember picture,overlay]
\node[anchor=south,yshift=10pt] at (current page.south) {\fbox{\parbox{\dimexpr\textwidth-\fboxsep-\fboxrule\relax}{\copyrighttext}}};
\end{tikzpicture}%
}

\newcommand{\Tau}{\mathrm{T}}
\newcommand{\mathd}{\mathrm{d}}
\newcommand{\tmem}[1]{{\em #1\/}}
\newcommand{\tmmathbf}[1]{\ensuremath{\boldsymbol{#1}}}
\newcommand{\tmop}[1]{\ensuremath{\operatorname{#1}}}

\newcommand{\tmstrong}[1]{\textbf{#1}}

{\theorembodyfont{\rmfamily}}

\title{\LARGE \bf
Robust Indoor Localization with Ranging-IMU Fusion
}

\author{
  Fan Jiang$^{1,2,*}$, David Caruso$^{1}$, Ashutosh Dhekne$^{2}$, Qi Qu$^{1}$, Jakob Julian Engel$^{1}$, Jing Dong$^{1}$
\thanks{$^{*}$This work was mostly done during the internship at Meta.}
\thanks{$^{1}$Meta Reality Labs Research.
        {\tt\footnotesize \{dcaruso, qqu, jakob.engel, 
        jingdong\}@meta.com}}%
\thanks{$^{2}$Georgia Institute of Technology.
        {\tt\footnotesize \{fan.jiang, dhekne\}@gatech
        .edu}}%
}

\begin{document}

\maketitle
\copyrightnotice
\thispagestyle{empty}
\pagestyle{empty}
\begin{abstract}
Indoor wireless ranging localization is a promising approach for low-power and high-accuracy localization of wearable devices. A primary challenge in this domain stems from non-line of sight propagation of radio waves. This study tackles a fundamental issue in wireless ranging: the unpredictability of real-time multipath determination, especially in challenging conditions such as when there is no direct line of sight. We achieve this by fusing range measurements with inertial measurements obtained from a low cost Inertial Measurement Unit (IMU). For this purpose, we introduce a novel asymmetric noise model crafted specifically for non-Gaussian multipath disturbances. Additionally, we present a novel Levenberg-Marquardt (LM)-family trust-region adaptation of the iSAM2 fusion algorithm, which is optimized for robust performance for our ranging-IMU fusion problem.
We evaluate our solution in a densely occupied real office environment. Our proposed solution can achieve temporally consistent localization with an average absolute accuracy of $\sim$0.3m in real-world settings.
Furthermore, our results indicate that we can achieve comparable accuracy even with infrequent range measurements down to 1Hz.
\end{abstract}

\section{Introduction}

Indoor localization is a core requirement for virtual reality/augmented reality (VR/AR) devices and robots. 
Traditionally, accurate indoor 6 Degrees-of-Freedom (DoF) localization is performed using Visual-Inertial Odometry (VIO)~\cite{Mourikis07icra}, and visual Simultaneous Localization and Mapping (SLAM)~\cite{Leutenegger13rss,Qin18tro,MurArtal15tro}, in which camera images are the main source of information. 
However, exposing raw images of the surrounding environment to localization algorithms has privacy implications, and capturing and processing a large amount of raw image pixels lead to significant energy consumption and heat dissipation, which is not ideal for small form factor wearable devices.

In recent years, wireless ranging based indoor localization has attracted
significant research attention, due to its good accuracy and low power consumption.
For example, the average ranging power of an Ultra-Wide Band (UWB) system could be as low as 0.14mW~\cite{Polonelli22sas}, assuming 1Hz ranging frequency (2.5ms per ranging, 55mW average power during ranging), which fits well within the power envelope of wearable devices.

While promising, wireless ranging is not free of issues. In particular, non-line of sight (NLOS) propagation of wireless signals, where radios waves undergo reflections from surrounding objects, can deteriorate ranging accuracy. In indoor environments cluttered with objects, NLOS measurements can become dominant, making conventional Gaussian noise models ill-suited. Even more complex scenarios arise when a user lacks line-of-sight to any localization transceivers, a situation frequently encountered in multi-room setups.

\begin{figure}
    \centering
    \resizebox{1.00\columnwidth}{!}{\includegraphics{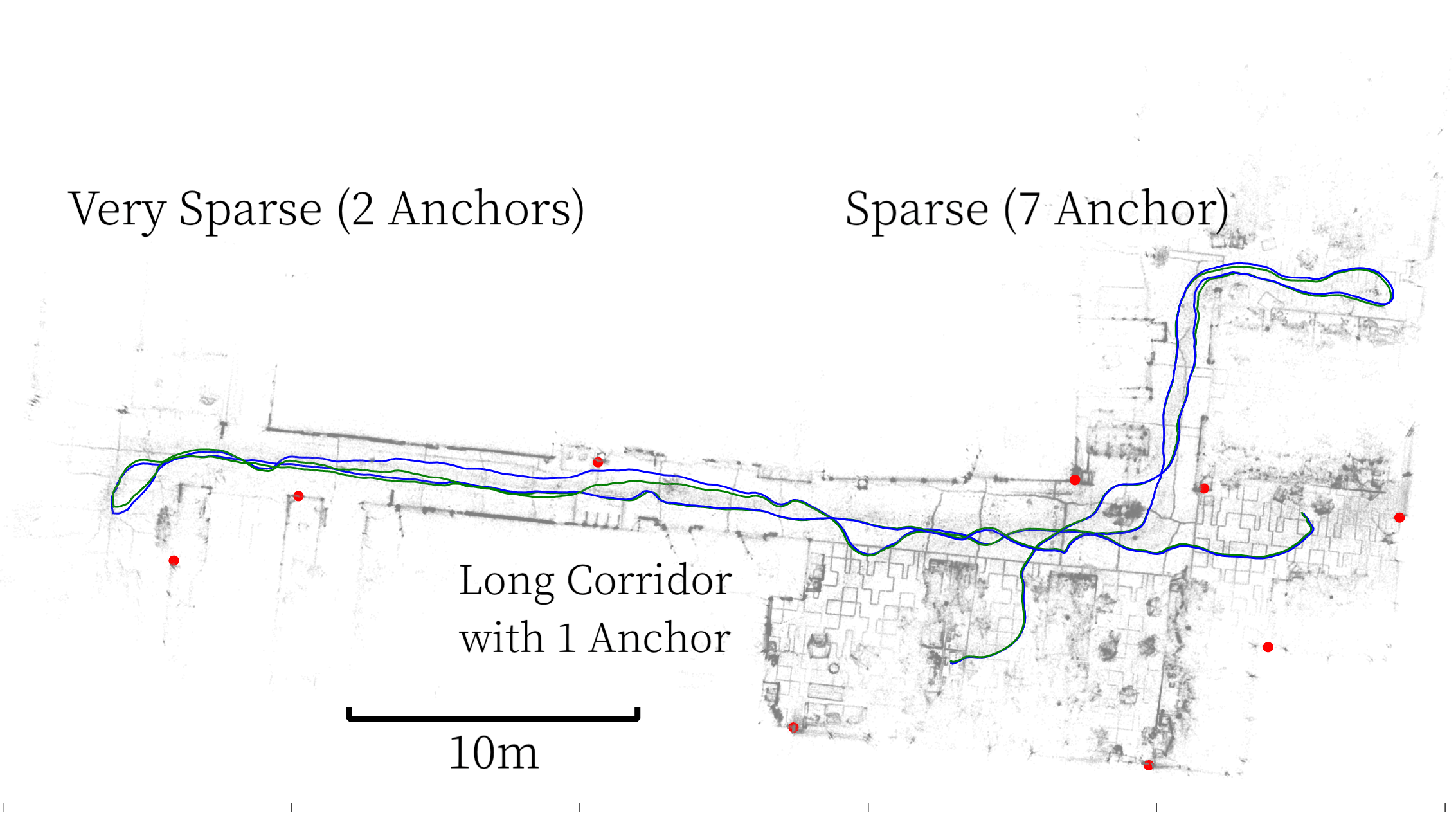}}
    \vspace{-0.8cm}
    \caption{Localizing in a cluttered, real office environment with sparsely populated anchors using proposed approach, with only UWB ranging and IMU sensor data. The average/max trajectory error are 0.3/1.0 meters. {\color{green}Green}: ground truth trajectory. {\color{blue}Blue}: result trajectory. {\color{red}Red}: UWB anchors. Note while the solution quality degrades in the corridor, due to very few line-of-sight ranging measurements in the corridor, it is able to recover once enough measurements are available.}
    \label{fig:indoor-result}
\end{figure}
Various hardware and signal processing methodologies have been proposed by the wireless community to mitigate multipath in ranging measurements. Notable strategies include using Channel Impulse Response (CIR)~\cite{Jiokeng20infocom}, multiple antennas (Multi-Input-Multi-Output, MIMO)~\cite{Zhao21imwut}, beamforming \cite{Roderick06ijss} and angle-of-arrival~\cite{Zhao21imwut}. 
On the other hand, the state estimation community tries to address the same problem by fusing together wireless ranging measurements with other sensors, to reduce ambiguity in multi-path determination during measurements~\cite{Qin23sensors,Corrales08acmhri,Cossette21ral}.

Depending on the current estimate's uncertainty, sufficient data may not be available for identifying NLOS measurement causally at all time, so the system needs to keep a long horizon of previous nonlinear information and correct past states as necessary. Such a requirement aligns seamlessly with incremental smoothing techniques like~\cite{Kaess12ijrr}.

In this work, we present a novel solution to the ranging-IMU fusion problem.
Our main contributions are:
\begin{enumerate}

  \item Designed, built and evaluated a system that can robustly localize a custom device composed of an IMU and UWB receiver in real-time, even in challenging environments. 
 
  \item A novel asymmetric $m$-Estimator for wireless Time-of-Flight (ToF) ranging measurement, to correctly model non-Gaussian multi-path effects;
  
  \item A practical way to improve the numerical stability of iSAM2, with presence of strong local non-linearity, without using
  complex trust region update strategies like Dogleg;
  
\end{enumerate}

We demonstrate that even with single-sided two-way ranging (SS-TWR) and sparsely placed UWB anchors we can achieve consistent indoor localization with range measurements with a mean absolute accuracy of $\sim$0.3m, and maintain reasonable location accuracy even with 1Hz measurements (see Figure~\ref{fig:indoor-result}).

The rest of the paper is organized as follows: we discuss related work in Sec.~\ref{sec:related_work} followed by the model used for the sensors in Sec.~\ref{sec:range_model}. In particular, it includes the model used for wireless ranging measurements: the main contribution of the paper. The improvement of the robustness of iSAM2, is described in Sec.~\ref{sec:isam2}, which forms the second contribution of this work. We finally discuss the implementation details of our prototype and our evaluation results in Sec.~\ref{sec:evaluation}. 

\section{Related Work}\label{sec:related_work}

\subsection{Related Work in Wireless Signal Processing}

Many methods have been proposed to reduce the effect to multipath and improve
the probability of detecting the ``first path" (the shortest propagation path) in complex indoor environments, for
example with neural networks~\cite{Tran22iros} and with signal processing techniques~\cite{Kolakowski17telfor}, sometimes
even with multiple antennae (Multi-Input-Multi-Output, MIMO)~\cite{Zhao21imwut}. 
A notable strategy utilizes CIR of received signals to
discern the direct-path measurements~\cite{Jiokeng20infocom, cao20206fit}, where each ``peak'' in the energy part of the CIR corresponds to a different
propagation path of the original transmitted signal in the space between the
transmitter and the receiver. The idea is to use the CIR to identify the first path in environments with
complex multipath effects. However, this approach is not robust in real world applications since path resolution below the limitations imposed by the wireless signal's bandwidth is challenging. The CIR often provides ambiguous information without knowing the detailed environment structure.
For example, when the direct path is blocked by walls, or has very low signal strength, any path identified by these approaches will lead to wrong range measurements.
Efforts using signal processing to identify such edge cases can lead to excessive computation and diminishing returns~\cite{Kong20sensors}.
While techniques like beamforming \cite{Roderick06ijss} and angle-of-arrival~\cite{Zhao21imwut}
provide supplementary insights, they often need additional hardware, and yet the fundamental challenges remain.

\subsection{Related Work in State Estimation}

Existing work try to solve multipath problems with sensor fusion, where the ranging
measurements are fused with a interoceptive motion sensor, usually an Inertial
Measurement Unit (IMU). Some of the first works in this space use Extended Kalman Filter (EKF) and prediction error based outlier rejection~(\cite{Qin23sensors,Corrales08acmhri,Cossette21ral}). After the proposal of the pre-integration technique \cite{Lupton12tro}, researchers have used smoothing and mapping techniques, for example factor graph optimization with iSAM2~\cite{Stromberg17thesis}. 
However, filtering approaches using IMU fusion without utilizing past data can only identify sparse outliers in range measurements. Such techniques cannot recover when valid ranging measurements are absent for significant time, since prediction is likely to drift away, worsening the problem for future valid measurements.

One possible way of modeling the effect of
non-line of sight measurement is to capture the inlier/outlier ambiguity
explicitly by using binary discrete variables in a {\tmem{hybrid}} factor graph. 
When incremental inference is required, explicit incremental solvers such as iMHS
{\cite{Jiang21arxiv}}, NF-iSAM {\cite{Huang21icra}}, or MH-iSAM2
{\cite{Hsiao19icra}} can be used. However, such hybrid modeling require mixed-integer program (MIP) solvers or
Expectation-Maximization (EM) style solvers~{\cite{Doherty22ral}}, whose cost could be prohibitive in
real-time applications.

Another common way to model outliers is {\tmem{m}}-Estimator.
$m$-Estimators have already been shown to map directly to E-M methods {\cite{Lee13iros}} in the continuous formulation.
The {\tmem{m}}-Estimator
for Cauchy distributions can be attributed to Barnett's 1966 work {\cite{Barnett66biometrika}}. 
The frequently referenced article on
$m$-Estimators {\cite{Zhang97ivc}}, makes an assertion---without specific
citation---that the Cauchy $m$-Estimator often produces incorrect results
without a means of verifying their accuracy. While this assertion holds merit,
it somewhat oversimplifies Barnett's original proposition.
Authors of {\cite{Barnett66biometrika}} specifically postulated that any local approach
resembling Newtonian methods has the possibility of being unsuccessful, however
such instances might be infrequent in practice.
Moreover, he emphasized that
evading local minima is unfeasible without a comprehensive exploration of the
likelihood function---a statement universally applicable to all
gradient-driven local methodologies.
Our proposed approach is based on $m$-Estimators, and we will demonstrate with our modified half-Cauchy $m$-Estimator, our approach delivers promising results.

\section{Ranging-IMU Measurement Model}\label{sec:range_model}

\subsection{IMU Model}
\label{sec:imu_model}
We use a typical IMU model. Assuming zero noise and known initial condition and a flat earth approximation, the simplified strapdown mechanization equations are given by:
\begin{eqnarray}
  \frac{\mathd}{\mathd t} \tmmathbf{R}_t & = & \tmmathbf{R}_t 
  [\tmmathbf{\omega}_t]_{\times} \nonumber\\
  \frac{\mathd}{\mathd t} \tmmathbf{V}_t & = & \tmmathbf{g}+\tmmathbf{R}_t
  \tmmathbf{a}_t \label{eq:rigid-dyn} \\
  \frac{\mathd}{\mathd t} \tmmathbf{X}_t & = & \tmmathbf{V}_t \nonumber
\end{eqnarray}
where
$\tmmathbf{R}_t$ is the rotation matrix, $\tmmathbf{V}_t$ the body velocity in
earth frame, $\tmmathbf{X}_t$ the translation vector, $\tmmathbf{g}$ the
gravity vector in the earth frame, $\tmmathbf{\omega}_t$ is current angular velocity in the IMU frame as measured by the gyroscope, and
$\tmmathbf{a}_t$ the current non-gravitational acceleration in the IMU frame as measured by the accelerometer. 
After applying a factory calibration to the IMU signal, we assume the rectified signal to be polluted by a Gaussian noise and a slowly varying biases, 3 components for the accelerometer bias $\tmmathbf{b}^a_t$ and gyroscope axis $\tmmathbf{b}^g_t$ . To model the later we chose a stochastic  1st-order Gauss-Markov random processes described here \cite{Carpenter18tr} (Section 5.2.4):
\begin{equation}
\frac{\mathd}{\mathd t}\tmmathbf{b}_t = - \dfrac{1}{\tau_{\text{bias}}} \tmmathbf{b}_t + w_{\text{bias}}(t)
\end{equation}
Where $\tmmathbf{b}_t = [\tmmathbf{b}^a_t, \tmmathbf{b}^g_t]^{\Tau}$ denotes both bias components of the IMU. $\tau_{\text{bias}}$ are a correlation time constant and $w_{\text{bias}}$ random variables following a centered Gaussian distribution with (diagonal) covariance $\Sigma_{w_{\text{bias}}}$.
The parameters of the model were derived from a recorded Allan Variance method slightly inflated to increase robustness to unmodeled effects.

This IMU model is leveraged with two types of factor in our optimization based smoother. The propagation constraint is using the {\tmem{preintegration}} technique as described in {{\cite{Lupton12tro}}, \cite{Forster15rss}}. The latter allows to easily write the likelihood of the preintegration measurement as:
\begin{multline}
    L_\text{preint} = \\
\exp\left(-\frac{1}{2} \left\| r_{\text{IMU}}\left((\tmmathbf{R}\tmmathbf{V}\tmmathbf{X})_{t+1}, (\tmmathbf{R}\tmmathbf{V}\tmmathbf{X})_{t}, \tmmathbf{b}_t\right) \right\| ^2_{\Sigma_{\text{IMU}}}\right)
\end{multline}
Where the residual $r_{\text{IMU}}$ and its covariance $\Sigma_{\text{IMU}}$ are defined respectively by Eq.~37 and Eq.~35 in \cite{Forster15rss} substituting $i$ by $t$ and $j$ by $t+1$.
The biases constraints are written with a binary factor  between consecutive bias estimates which represents the likelihood:
\begin{equation}
   L_{\text{bias}} = \exp\left(-\frac{1}{2} \left\|\tmmathbf{b}_{t+1} - e^{\frac{\Delta t_{t;t+1}}{\tau}}\tmmathbf{b}_{t} \right\|^2_{\Sigma_{w_{\text{bias}}}}\right)
\end{equation}

\subsection{Ranging Measurement Model}

The wireless ranging measurement factor's likelihood is modeled with a mixture of two
probability distributions. One distribution is for the line of sight measurements, where
the noise mainly comes from the inaccuracies of the First Path Estimation (FPE)
algorithm. This error is mainly caused by inaccuracies in the radio hardware,
as well as the limited energy and bandwidth of the transmitted radio signal.
This noise appears in the range measurement as a Gaussian-like noise of a
$\sigma$ of about 0.1-0.2m~\cite{Schuh19thesis}. 
The other distribution is for the NLOS measurements, where the measurements
follow an {\tmstrong{environment-dependent}} unknown distribution, which could
be as large as twice of the actual distance, or as small as indistinguishable
from the Gaussian measurement noise.
\begin{equation}
  L_{\text{range}} (r| \bar{r}, m) \sim \left\{\begin{array}{ll}
    \mathcal{N} (\bar{r} ; \sigma_r), & m = 0\\
    L_{\text{range}} (r| \bar{r}, m = 1), & m = 1
  \end{array}\right.
\end{equation}
where $r$ is the measured range $r=\left\| \tmmathbf{X}_i - \tmmathbf{A}_j \right\|_2$ where $\tmmathbf{A}_j$ is the anchor $j$'s 3D position, $\bar{r}$ is the true range, $\sigma_r$ is the
range variance, and $m = \{ 0, 1 \}$ is the discrete variable that indicates if
the measurement is a direct path measurement or not, with $m = 0$ indicates the range measurement is direct, while $m = 1$ indicates it is not, which we will dwell on next.

Existing work~\cite{Rosen13icra} factorizes the NLOS distribution
into two distributions with a convolution of the Gaussian and an exponential distribution, which is the maximum entropy distribution supported on $[0, +
\infty)$ with some statistical moment, with a sample mean and variance
obtained by simulation of the ranging process in a simulator.
However, in wireless ranging problems we cannot apply the same model as in {\cite{Rosen13icra}},  since the environment is unknown and hence we have little information about $p (r| \bar{r}, m =
1)$, 

Instead, as the first contribution of this paper, we model the NLOS measurements with a half-Cauchy distribution, in line with the maximum entropy principle, with location $\theta = 0$
and scale $\gamma$, supported also on $[0, + \infty)$:
\begin{equation}
  p (r' - \bar{r} |m = 1) \sim \frac{2}{\pi \gamma}  \frac{1}{1 + ((r'-r) /
  \gamma)^2}
\end{equation}
where $r'$ is the measured range, $\bar{r}$ the true
range.
The unknown scale parameter $\gamma$ can be either determined by parameter
estimation techniques using real or simulated data, or simply derived using
the Interquartile Range of the (Gaussian) ranging noise heuristically.

\begin{figure}[ht]
  \resizebox{0.5\columnwidth}{!}{\includegraphics{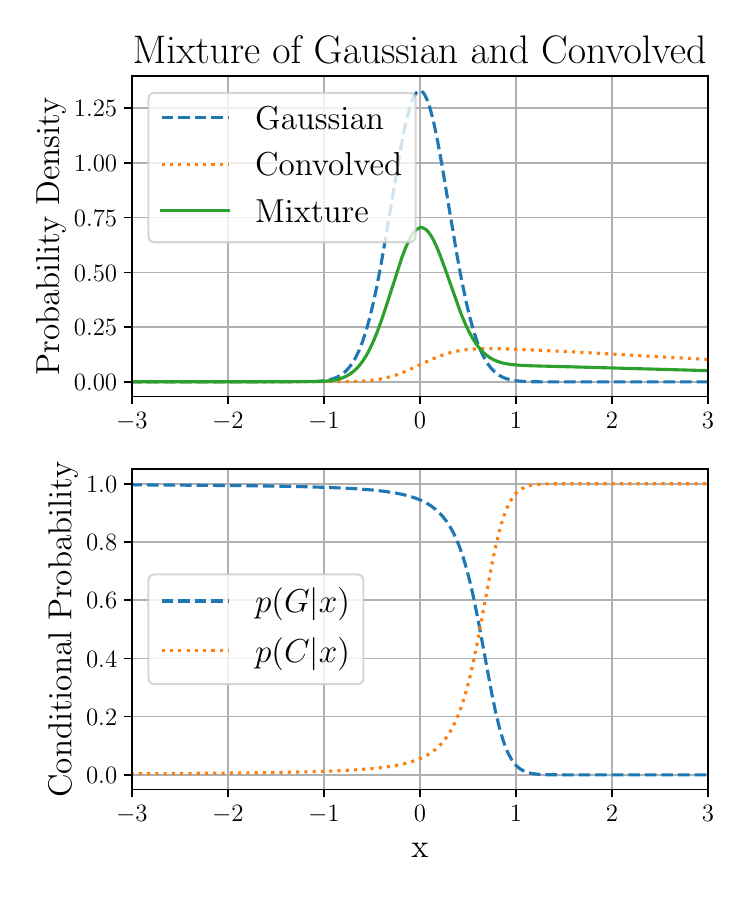}}\resizebox{0.5\columnwidth}{!}{\includegraphics{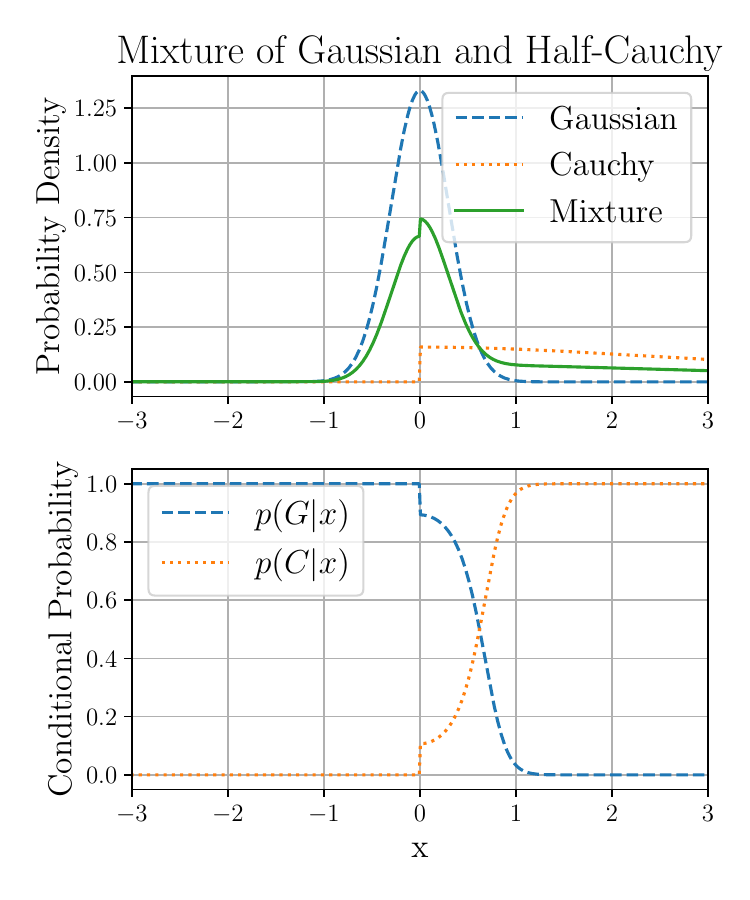}}
  \caption{Qualitative comparison of using (a) a combined marginal model and
  (b) a mathematically simpler half-Cauchy model for the multipath. Note the less complex
  model does not significantly change the decision
  point.\label{fig:prob-conv-truncated}}
\end{figure}

With this NLOS model, the observed range model for $m = 1$ is then (sum
of independent variables is convolution)
\begin{equation}
  L_{\text{range}} (r| \bar{r}, m = 1) = \int_{- \infty}^{+ \infty} p(r|r') p(r' - \bar{r}
  |m = 1) \mathd r'
\end{equation}
which is shown in Fig. \ref{fig:prob-conv-truncated}a.

The physical intuition behind the use of a single-sided distribution for NLOS measurements, is that the true range is the shortest path in the space, and NLOS measurements are necessarily longer than the shortest path between anchor and receiver. 

\begin{figure}
    \centering
    \resizebox{0.9\columnwidth}{!}{\includegraphics{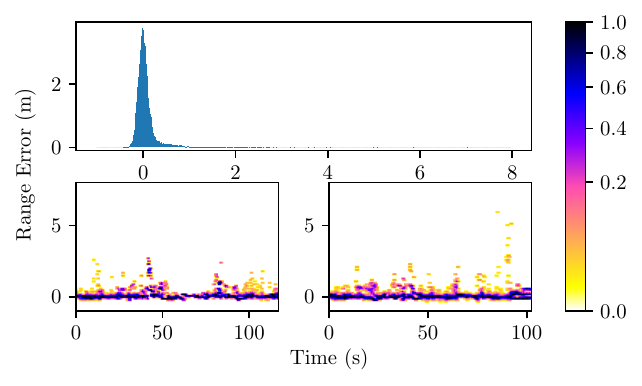}}
    \caption{Range measurement error distribution. Top: marginal over all trajectories seems light-tailed. Bottom: heat-map over time for run 1 and 2 of Table \ref{tbl:ape_40hz}. Note distribution becomes fat tailed in a lot of places, for example $t\sim 45s$ on the left, and $t\sim 80s$ on the right. Color indicates the value of the probability density function at time $t$.}
    \label{fig:range-error-time}
\end{figure}

In Fig.~\ref{fig:range-error-time} we show the evolution of the UWB ranging measurement error distribution across trajectories, using the ground-truth ranges computed through ground truth trajectories and anchor positions. 
The x axis is time elapsed, y axis is the range error in meters, and the color represents the distribution density. It is apparent that the distribution is single-sided, with occasional large outliers, and frequent smaller outliers.

It is important to note that while the marginal distribution for $r-\bar{r}$ may seem light-tailed, the actual distribution considering the hidden variable, the current surroundings of the user, is not. This can be observed on Fig. \ref{fig:range-error-time}, where persistent NLOS measurements dominate in some time slices. This explains why we need to use a half-Cauchy distribution without priors on $m$, instead of just using the measured range marginal density like in \cite{Rosen13icra}.

Other than considering explicit hybrid factor graph inferences to solve discrete variable $m$, 
we use an \emph{implicit} approach to solve the range-IMU fusion problems, 
based on the $m$-Estimators to simplify the inference process. 
Since the decision boundary (where $m = 0$ or $m = 1$ is more likely) is the same, and the p.d.f. of both the marginal and the half-Cauchy distribution is very close after the decision boundary, we can use the half-Cauchy distribution directly as the mixture component (Fig. \ref{fig:prob-conv-truncated}b). 

\begin{figure}
    \centering
    \resizebox{0.95\columnwidth}{!}{\includegraphics{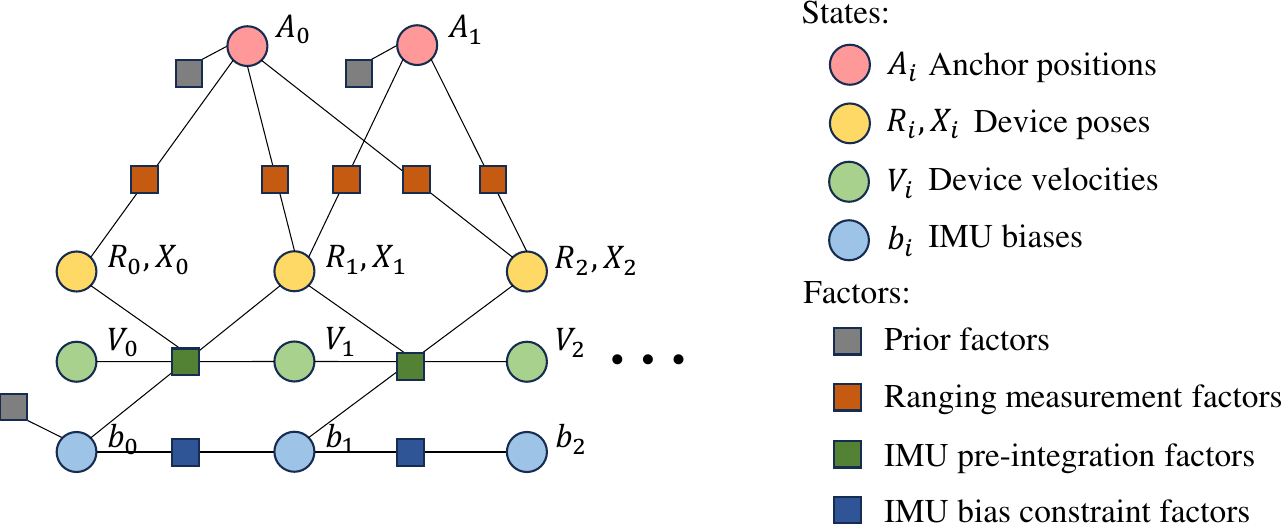}}
    \caption{An example factor graph of the ranging-IMU fusion.}
    \label{fig:factor-graph}
\end{figure}

The IMU and ranging model presented are used in a factor graph formulation. An example factor graph is shown in Fig.~\ref{fig:factor-graph}.
The states include the devices pose + velocity + IMU bias at each time point when we receive a range measurement, and also positions of all rang-able anchors. The factors include the IMU pre-integration factors and ranging factors we just introduced, 
plus prior factors on first IMU bias with factory calibration, 
and all anchors with pre-mapped positions. 

\section{Trust-Region Variant of iSAM2}\label{sec:isam2}

A known issue of using iSAM2 framework in incremental inference is
the occurrence of indeterminant linear systems {\cite{Huai21aaai}}. This is
primarily because iSAM2 internally uses a Gauss-Newton like
update for solving the nonlinear least-squares problem, which is not robust to ill-conditioned problems.

While existing trust-region based methods like RISE {\cite{Rosen14tro}} have been shown
to perform well in regular SLAM problems, its convergence has not been
validated with switchable methods like ours whose continuous error could
change with the discrete decision variable. Same as previously reported~\cite{McGann23icra},
we observed that RISE does not work when the radius of the trust region changes. This leads to a
performance similar to Gauss-Newton (vanilla iSAM2), which similarly fails to achieve optimal non-linear updates, resulting in worse performance in our application.

In contrast to the Dogleg-like algorithm proposed by {\cite{McGann23icra}}
(which is concurrent to our work), as the second contribution of this paper, we propose {\tmstrong{D-iSAM2}}, a simpler trust-region
method, which works well in our real-world experiments, and only requires minimal changes in
iSAM2. This method shares the same core idea with the Levenberg-Marquadt
(L-M) algorithm, whose linear update (in the simplest form) is calculated from
\begin{equation}
  (J^{\Tau} J + \lambda I) \delta = J^{\Tau} z
\end{equation}
where $J$ is the Jacobian, $\lambda$ the damping factor, $\delta$ the linear
increment, $z$ the linear error vector. This linear problem effectively is equivalent to solving a nonlinear
factor graph with the following Jacobian structure
\begin{equation}
  \left[\begin{array}{cc}
    J^{\Tau} & \lambda^{1 / 2} I
  \end{array}\right]  \left[\begin{array}{c}
    J\\
    \lambda^{1 / 2} I
  \end{array}\right] \delta = \left[\begin{array}{cc}
    J & \lambda^{1 / 2} I
  \end{array}\right]  \left[\begin{array}{c}
    z\\
    0
  \end{array}\right]
\end{equation}
which is equivalent to the original graph with a {\tmem{special}} factor on
each variable where the factor always has an error function of value $0$, but
a Jacobian of $\lambda^{1 / 2} I$. When the trust region is an ellipse, $\lambda I$ can be replaced by a diagonal of $n$ lambdas, $\Lambda = \mathrm{diag}(\lambda_1,\dots,\lambda_n)$.

When new information is added, the algorithm only needs to check whether the current
$\lambda$ is valid, by checking if the error decreases after the iSAM2 step with the special factor added. If
not, increase $\lambda$, and if yes, decrease $\lambda$, then repeat the process. We refer the interested reader to \cite{Nocedal06book} where the schedule for the $\lambda$ modifications have been extensively covered.

\makeatletter
\renewcommand{\ALG@beginalgorithmic}{\small}
\makeatother

\begin{algorithm}[h]
\caption{D-iSAM2 Algorithm}\label{alg:d-isam2}
\renewcommand{\algorithmicrequire}{\textbf{Input:}}
\renewcommand{\algorithmicensure}{\textbf{Output:}}
\begin{algorithmic}
\Require Bayes Tree $\mathcal{T}$, current estimate $\mathcal{X}$, new factors $\mathcal{F}$$=\{\phi_i\}$, new variable initial estimates $\{x_i\}$, current $\lambda$
\State convergence indicator $c \gets$ False
\While{not converged $c$}
\State $\mathcal{F}' \gets$ AddSpecialFactor($\mathcal{F}$, $\lambda$)
\State $\mathcal{T}',\mathcal{X}' \gets$ iSAM2Update($\mathcal{T}, \mathcal{F}', \mathcal{X}, \{x_i\}$)
\If{nonlinear error decreased}
    \State $c \gets $ True
    \State Decrease $\lambda$ with schedule  \Comment{until $\lambda_{\text{min}}$}
\Else
    \State Increase $\lambda$ with schedule
    \State if $\lambda=\lambda_\text{max}$ throw error
\EndIf
\EndWhile
\Ensure new Bayes Tree $\mathcal{T}'$, new estimate $\mathcal{X}'$
\end{algorithmic}
\end{algorithm}

\begin{figure*}[ht]
  \centering\resizebox{17.5cm}{!}{\includegraphics{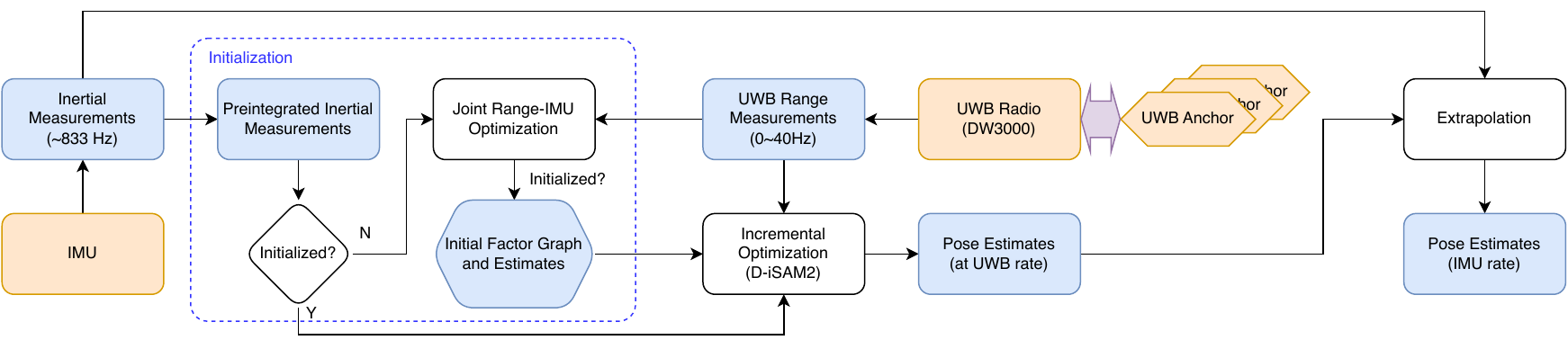}}
  \vspace{-0.3cm}
  \caption{System setup of our inertial-ranging fusion system. White blocks are algorithm, yellow blocks are hardware, and blue blocks are data.}\label{fig:system_diag} 
\end{figure*}

Note that the $\lambda_{x_i}$ for an old variable $x_i$ will not change after a successful D-iSAM2 step. This is intentional, since it is very costly to update all old Bayes tree nodes for a new set of $\{\lambda_{x_n}\}$, since all nodes will need to be re-factorized. However, since we operate incrementally, the last added variable is the most numerically challenging variable. Hence, only changing the trust region radius for the last variable is sufficient.
The whole algorithm is described in Alg. \ref{alg:d-isam2}.

Changing the $\lambda$ for past variables, which we explicitly choose not to do, is a ``global'' process similar to full
relinearization in iSAM2. The process impacts every node in the factor graph, hence could be detrimental to the overall performance if done in a naive way. However, this may be required in some problems more difficult than ours. Also, child
nodes can be marginalized once they are too old for the current estimate, like
in a fixed lag smoother. Since this is not the primary aim of this paper, we
will leave how to further optimize this trust-region based robust incremental
optimization method for a future work.

\section{Implementation}\label{sec:implementation}

The overall system diagram is shown in Figure~\ref{fig:system_diag}. Input sensor data of the system include an IMU operated at high frequency, and a UWB transceiver that executes SS-TWR range measurements from fixed anchors in the space. During the initialization phase, a joint of factor graph of ranging data and preintegration of the IMU data is built for 10 seconds, and a batch optimization of the factor graph is performed to obtain initial values of the system states, to initialize an iSAM2 estimator. After initialization, the raw ranging measurements and preintegrations are directly fed into iSAM2 to get UWB-frequency estimation, and IMU-frequency estimation can be future obtained by extrapolation using IMU measurements.

\begin{figure}[t]
\centering
\resizebox{0.8\columnwidth}{!}{\includegraphics{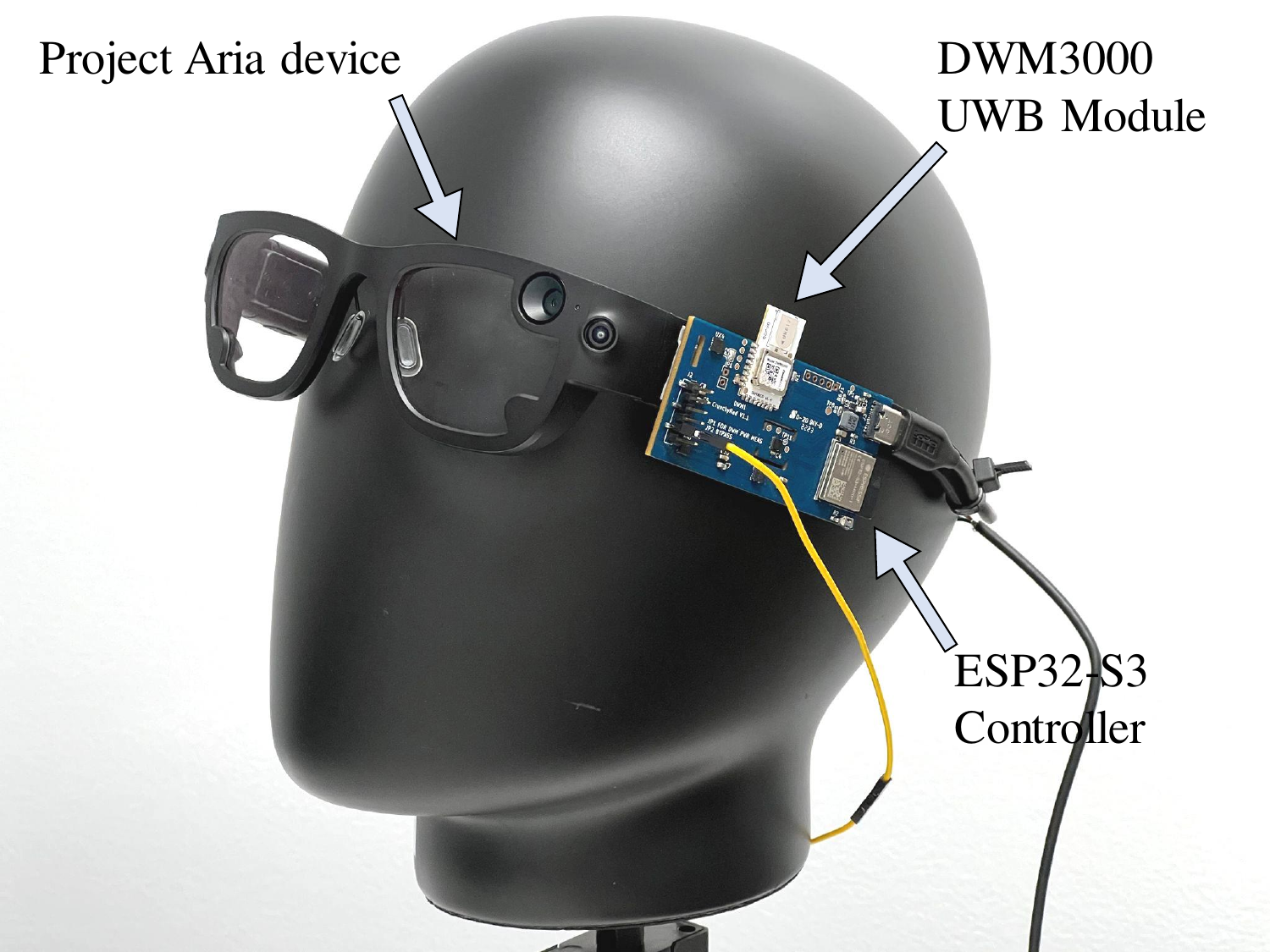}}
\caption{Prototype hardware built, with a wireless module attached to a Project Aria
device~\cite{Somasundaram23arxiv}.}\label{fig:hardware}
\end{figure}

\label{sec:impl}We implemented the wireless hardware using the Decawave DWM3000
module with ESP32-S3 as the main microcontroller. For the ranging protocol, we
use a simple SS-TWR ranging scheme, where the device sends 1 packet to the
anchor being ranged against and receive 1 packet with timestamp. We use SS-TWR because it possesses similar noise characteristics with Wi-Fi Fine Time Measurement (FTM), which enables our method to also apply to Wi-Fi localization.
The IMU stream is directly recorded off the left IMU of the Project Aria
device~\cite{Somasundaram23arxiv}, which is a factory calibrated BMI263 from Bosch operating at 800Hz.
The IMU stream is time synchronized with the UWB data, through a hardware time synchronization link between the UWB module and Project Aria device.

The fusion algorithm is implemented with C++ using the GTSAM
{\cite{Dellaert17fnt}} library using the preintegrated IMU factor
of Sec.~\ref{sec:imu_model}. All evaluations are conducted
on a Macbook Pro with an Apple M1 Pro chip.

\section{Evaluation}\label{sec:evaluation}

We evaluate the proposed system via experiments in a typical 30 by 50 meters office environment.
Ground truth device trajectory is
obtained with Project Aria Machine Perception Service~\cite{Somasundaram23arxiv}, using Project Aria device's collected data as input. UWB anchors are co-located in the the ground truth trajectory frame of reference using 2D
fiducials, with $< 1 \tmop{cm}$ accuracy.
We evaluate the performance of our approach over 7 runs, each of which is a walk
1-3 minutes in duration and 50-130 meters long in travelled distance.
The resulting trajectory is evaluated using the Evo library
{\cite{Grupp17evo}}.

Ranging frequency of each anchor is set at maximum 10Hz, and the receiver is configured to output measurements from up to 4 UWB anchors which have strongest signal strength, so the cumulative received ranging measurements are up to 40Hz. We will use 40Hz ranging frequency in Section~\ref{sec:evaluation_40hz}, and explore lower frequency in \ref{sec:evaluation_1hz}.

\begin{table*}[t!]
  \centering
  \vspace{0.2cm}
    \begin{tabular}{cc|ccccccc}
    \hline
      Noise model & Estimator & Run 1 & Run 2 & Run 3 & Run 4 & Run 5 & Run 6 & Run 7 \\ \hline 
        Gaussian & D-iSAM2 & 0.68/4.44 & 0.90/2.28 & 0.59/1.98 & 0.43/0.98 & 1.08/10.43 & \textbf{0.33}/\textbf{1.30} & 0.66/6.06 \\
        Huber & D-iSAM2 & 0.54/4.02 & 0.35/1.83 & 0.52/\textbf{1.15} & \textbf{0.30}/\textbf{1.31} & 0.45/5.05 & 0.70/2.57 & 0.41/1.31 \\
        Cauchy & D-iSAM2 & 0.36/1.27 & 0.78/1.62 & 0.62/1.74 & \textbf{0.30}/1.40 & 0.26/2.73 & 0.85/2.77 & 0.39/1.11 \\
        Proposed & D-iSAM2 &  \textbf{0.33}/\textbf{1.20} & 0.19/0.76 & \textbf{0.47}/1.37 & 0.33/1.54 & \textbf{0.18}/\textbf{1.15} & 0.59/2.35 & \textbf{0.30}/\textbf{1.07} \\
      \hline
    Proposed & RISE \cite{Rosen14tro} & \textbf{0.32}/\textbf{1.22} & \textbf{0.13}/\textbf{0.36} & 0.59/1.64 & 0.32/1.52 & 0.32/2.97 & 0.59/2.35 & \textbf{0.31}/\textbf{1.06}\\
      Proposed & iSAM2 & 0.70/2.27 & 0.35/0.94 & 0.77/2.40 & 0.62/2.64 & 0.32/3.46 & 0.60/2.95 & 0.50/1.78 \\
      Huber & iSAM2 & 1.00/5.56 & 0.51/2.29 & 0.84/1.90 & 0.64/2.34 & 0.57/7.17 & 0.79/3.14 & 0.61/1.79 \\
      \hline
    \end{tabular}
  \caption{3D absolute position error (meters) with different ranging noise model and estimators (average/max), using 40Hz range measurements. The best method is indicated in highlighted with 5\% tolerance.
  \label{tbl:ape_40hz}}
\end{table*}

\begin{table*}[t!]
  \begin{center}
    \begin{tabular}{c|ccccccc}
    \hline
      Noise model & Run 1 & Run 2 & Run 3 & Run 4 & Run 5 & Run 6 & Run 7 \\ \hline
      Gaussian & 1.81/8.49 & 1.07/2.99 & 0.68/3.30 & 0.73/2.42 & 2.90/15.67 & \textbf{0.91}/\textbf{3.97} & 1.06/3.68 \\
Huber & Fail & Fail & Fail & Fail & 1.03/5.73 & Fail & 1.07/2.35 \\
Proposed & \textbf{0.49}/\textbf{1.44} & \textbf{0.45}/\textbf{1.61} & \textbf{0.61}/\textbf{3.01} & \textbf{0.73}/\textbf{2.28} & \textbf{0.64}/\textbf{4.15} & 1.30/5.54 & \textbf{0.46}/\textbf{1.37} \\
\hline
    \end{tabular}
  \end{center}
  \caption{3D absolute position error (meters)  using 1Hz range measurements (average/max). If the average error is greater than 10 meters we classify the result as failure.\label{tbl:ape_1hz}
  }
\end{table*}

\subsection{Evaluation of proposed ranging model: 40Hz}\label{sec:evaluation_40hz}

We evaluate our proposed asymmetric model against Gaussian, also standard $m$-Estimators Cauchy and Huber, with the proposed D-iSAM2 estimator.
The resulting metrics are shown in Table \ref{tbl:ape_40hz}. Our implicit hybrid method with asymmetric model beats all baselines in most runs, in both average and maximum trajectory errors.

\subsection{Evaluation of proposed ranging model: lower frequency}\label{sec:evaluation_1hz}

\begin{figure}[h]
    \centering
    \centering\resizebox{0.7\columnwidth}{!}{\includegraphics{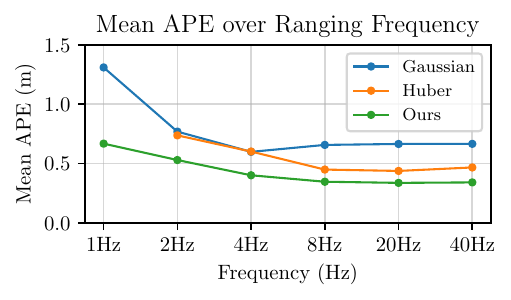}}
    \caption{Mean 3D APE over all datasets with different ranging frequency. Huber fails to give results on 1Hz (high APE due to estimation divergence).}
    \label{fig:ape-vs-freq}
\end{figure}

Reducing the ranging frequency will enable localization with lower power, at the cost of increased difficulty of outlier identification because open-loop IMU integration's accuracy degrades quickly over time.
We still use D-iSAM2 as incremental estimator, and keep all other parameters same as previous section.

In Table~\ref{tbl:ape_1hz} we pushed our evaluation to only using 1Hz ranging measurements. We find using Huber model causes complete estimation divergence on most of the sequences, due to Huber loss function identifies the true line-of-sight measurements as NLOS measurements. The Gaussian model, effectively treating all measurements as inliers,  does not fail, but also leads to higher trajectory error.
In Figure~\ref{fig:ape-vs-freq} we show the trajectory error results on a range of frequencies between 1Hz to 40Hz. We can find that our proposed model always have the best trajectory accuracy on all frequencies, and Huber is the second best except for its divergence observed at 1Hz.

\subsection{Evaluation of proposed D-iSAM2}\label{sec:evaluation_estimator}

\begin{table}[h]
    \centering
    \begin{tabular}{c|cc}
    \hline
        Estimator & Average Runtime (ms) & Max Runtime (ms) \\ \hline
        iSAM2 & 21.33 & 83.19 \\
        D-iSAM2 & 19.21 & 79.46 \\
        RISE \cite{Rosen14tro} & 66.85 & 201.21 \\
        \hline
    \end{tabular}
    \caption{Per iSAM2 update runtime comparison between estimators.}
    \label{tab:runtime-compare}
\end{table}

We also compare our proposed D-iSAM2 estimator against the RISE \cite{Rosen14tro} Dogleg optimizer.
All parameters remain the same except the type of optimizer used, and ranging frequency is set to 40Hz. 
Initial delta for the Dogleg algorithm is set at 0.1. 
The runtime of each estimator is shown in Table \ref{tab:runtime-compare} as well as the quality of the optimized solution in the lower part of Table \ref{tbl:ape_40hz}. 
While the solution quality of RISE in some sequences match these obtained by our algorithm, the runtime of RISE is significantly longer.
D-iSAM2 reaches similar runtime as vanilla iSAM2, while beating iSAM2 on trajectory accuracy for a large extent.

\section{Conclusion}

We demonstrate that our proposed asymmetric noise model handles real-world range
noises better than existing $m$-Estimators, with ranging measurement rate as low as 1Hz, which paves the way
forward to accurate, energy-efficient indoor navigation using wireless
ranging.
We also show that our novel trust-region based incremental solver
effectively handles the nonlinear range-IMU fusion problem. In most cases, we
achieve both better convergence and lower run time than RISE and
vanilla iSAM2. 
Finally, we show an end-to-end system which can localize accurately in indoor spaces using only IMU and UWB ranging measurements, with the hardware and software system we built, which has great potential in future low-power wearable devices.







\bibliographystyle{plain} 
\bibliography{references} 

\end{document}